\documentclass[10pt,twocolumn,letterpaper]{article}

\usepackage[pagenumbers]{cvpr} 

\definecolor{cvprblue}{rgb}{0.21,0.49,0.74}
\usepackage[pagebackref,breaklinks,colorlinks,allcolors=cvprblue]{hyperref}
\usepackage{multirow}
\usepackage[accsupp]{axessibility}  
\def\paperID{7} 
\def\confName{CVPR}
\def\confYear{2026}

\title{Novel Anomaly Detection Scenarios and Evaluation Metrics \\ to Address the Ambiguity in the Definition of Normal Samples}

\author{
Reiji Saito, 
Satoshi Kamiya, and
Kazuhiro Hotta\\
Meijo University, 
1-501 Shiogamaguchi, Tempaku-ku, 
Nagoya 468-8502, Japan\\
{\tt\small \{200442065, 180442042\}@ccalumni.meijo-u.ac.jp, kazuhotta@meijo-u.ac.jp}
}

\begin{document}
\maketitle
\begin{abstract}
In conventional anomaly detection, training data consist of only normal samples. However, in real-world scenarios, the definition of a “normal sample" is often ambiguous. For example, there are cases where a sample has small scratches or stains but is still acceptable for practical usage. On the other hand, higher precision is required when manufacturing equipment is upgraded. In such cases, normal samples may include small scratches, tiny dust particles, or a foreign object that we would prefer to classify as an anomaly. Such cases frequently occur in industrial settings, yet they have not been discussed until now. Thus, we propose novel scenarios and an evaluation metric to accommodate specification changes in real-world applications. Furthermore, to address the ambiguity of normal samples, we propose the \textbf{RePaste}, which enhances learning by re-pasting regions with high anomaly scores from the previous step into the input for the next step. On our scenarios using the MVTec AD benchmark, RePaste achieved the state-of-the-art performance with respect to the proposed evaluation metric, while maintaining high AUROC and PRO scores. Code: \url{https://github.com/ReijiSoftmaxSaito/Scenario}.
\end{abstract}    
\section{Introduction}
\label{sec:intro}

In industrial manufacturing, anomaly detection aims to identify product defects and accurately localize them, particularly small defects that are difficult to detect through manual inspection. Misidentification caused by human fatigue or inattention remains a challenge, underscoring the need for automated inspection systems.
Recently, anomaly detection methods trained only on normal images \cite{GLASS,effAD,recon,msflow,glad} have attracted increasing attention. In these approaches, models are trained exclusively on normal samples due to the scarcity and diversity of anomalous data, and then distinguish between normal and anomalous samples during inference.

\begin{figure}[t]
    \centering
    \includegraphics[width=1.0\linewidth]{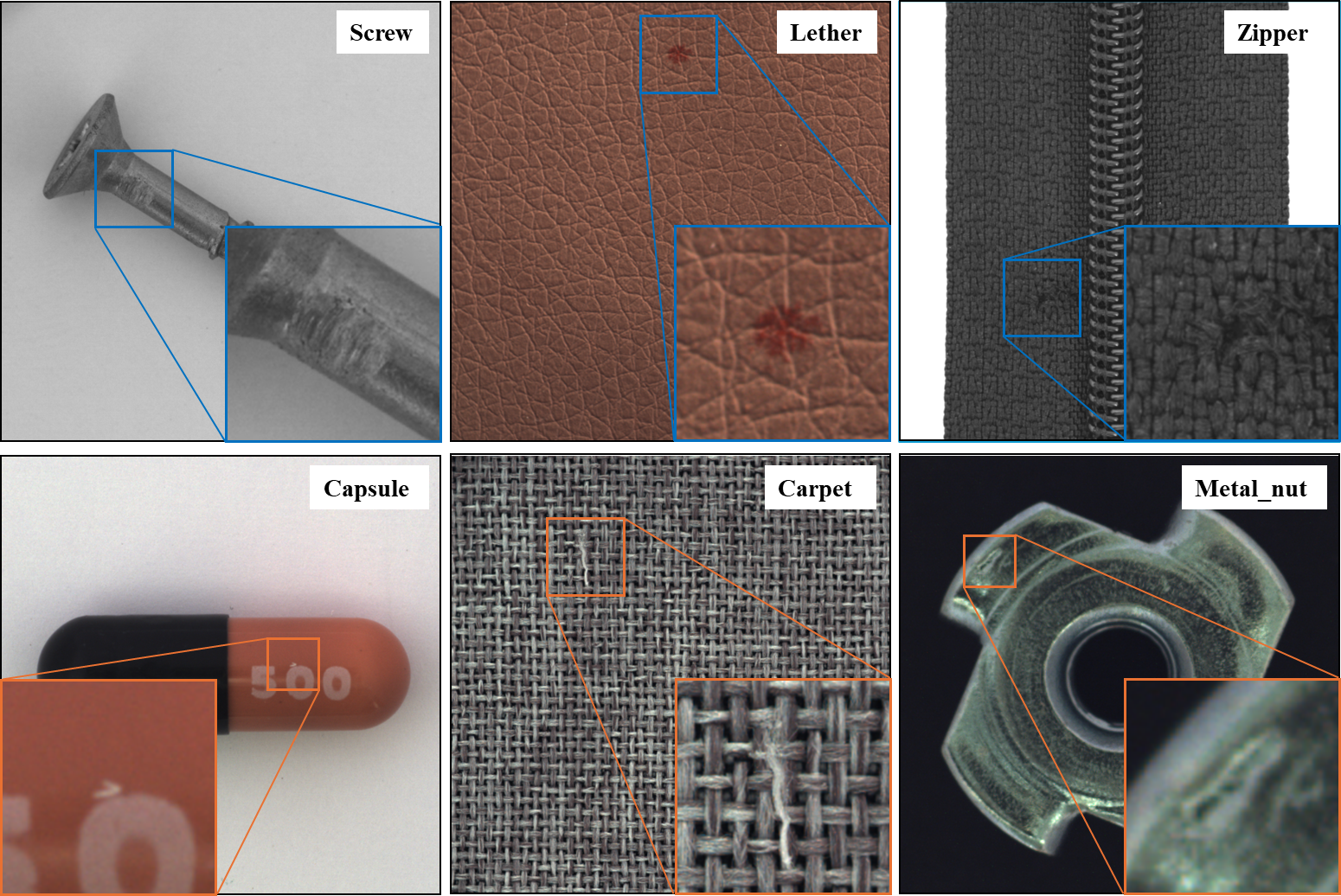}
    \caption{Examples that the definition of a normal sample is ambiguous. In the MVTec AD dataset, \textcolor{blue}{blue} boxes indicate anomalous samples, while \textcolor{orange}{orange} boxes indicate normal samples.
    }
    \label{fig:abst}
\end{figure}

In traditional anomaly detection, the training data include only normal samples without defects. However, in real-world use, the definition of “normal” is often unclear. For example, as shown in \cref{fig:abst}, small dust or tiny scratches in images may or may not be treated as normal, depending on the situation. In addition, the meaning of “normal” may change over time due to the change in the manufacturing environment or product design. Therefore, there are two situations to consider. (1) Samples once seen as an anomaly may later be considered as normal. (2) Normal samples may later be anomalous samples. Being able to adapt flexibly to these definition changes is very important in practice. Related research areas include concept drift \cite{ConceptDrift,survey}, domain adaptation \cite{uDA,DA_survey}, and continual learning \cite{continual,LwF}, which have been widely studied for handling changes in data distributions. However, these approaches primarily address distribution shifts and differ in nature from the problem setting considered in this study, namely, the explicit redefinition of anomaly and normal roles within the same visual domain.
To address these practical needs, this study proposes two novel scenarios: 
\begin{itemize}
    \item The “Anomaly-to-Normal scenario (A2N)”, in which anomalies such as cracks or glue are reclassified as normal due to specification changes.
    \item The “Normal-to-Anomaly scenario (N2A)”, in which normal samples are reclassified as anomalies due to specification changes.
\end{itemize} 
By using these scenarios, it becomes possible to discover models that can easily update the definition of normal samples, thereby expanding the range of practical applications.

To verify which method is effective in the proposed scenarios, it is important to use appropriate evaluation metrics for anomaly detection performance. 
In anomaly detection tasks, evaluation metrics such as AUROC \cite{auroc}, AUPRC, and F1-score are widely used to measure how accurately models distinguish between samples defined as normal or anomalous.
In the field of domain adaptation, performance degradation from a source domain to a target domain is often quantified using metrics such as Performance Drop Rate \cite{PDR}.
In continual learning, metrics such as Backward Transfer \cite{BT,forget} are employed to evaluate how the performance of previously learned tasks changes after learning new tasks.
However, these evaluation metrics assume that the semantic definition of labels remains unchanged. They are designed to assess discrimination performance under fixed anomaly criteria, robustness against distribution shifts, or knowledge retention across tasks.
In contrast, the setting considered in this study involves a redefinition of anomaly semantics. Specifically, samples that were previously regarded as anomalous may be reclassified as normal, and vice versa. Such changes in the definitions of normal and anomalous samples are fundamentally different from mere distribution shifts or the addition of new tasks. Rather, they require the reconstruction of the decision boundary itself.
Therefore, existing evaluation metrics cannot explicitly quantify a model’s adaptability to changes in anomaly definitions. To address this limitation, we introduce AUROC for Specification Changes (S-AUROC)
as a metric for measuring how well a model adapts to the redefinition of normal and anomalous samples caused by specification changes.
Unlike standard AUROC, S-AUROC explicitly focuses on the subset of samples affected by the redefinition and is used to compare models trained before and after specification changes, thereby quantifying adaptability.

To maintain high performance under the proposed two scenarios (A2N and N2A), an anomaly detection model that can flexibly adapt to changes in the definition of normal samples is required. 
To investigate this requirement, we conducted comparative experiments with many recent anomaly detection methods \cite{fastflow,patchcore,RD4AD,RDplus,simplenet,DiAD,mambaAD,INPFormer,UniNet,Dinomaly,GLASS}
under the proposed scenarios.
The results indicate that GLASS \cite{GLASS}, a pseudo-anomaly-based method, achieves the best performance among the comparison methods.
However, GLASS also has limitations. To flexibly handle specification changes, a model should be able to generate both transitions from normal to anomalous samples and from anomalous to normal samples. Since GLASS is based on pseudo-anomaly generation, it can only generate anomalies from normal samples, making it insufficient for fully addressing both types of specification changes. Due to this limitation, specification changes may alter the definition of normal samples, such that regions previously regarded as anomalous need to be treated as normal. Therefore, it is necessary to explicitly enable the model to re-learn such regions as normal samples.

To achieve this, we propose a method called \textbf{RePaste}, which enhances the learning by re-pasting regions with high anomaly scores at the previous step to the next image. Regions with high anomaly scores in training images often contain unusual features, such as small scratches, as shown in \cref{fig:abst}. 
Regions that consistently receive high anomaly scores often correspond to features that are rare in the training data. By re-pasting these regions to increase their frequency, the model is encouraged to adjust its decision boundary under redefined normal semantics, thereby improving its adaptability to changes in the definition of normal samples.

In the experiments, the proposed scenarios were evaluated on the industrial anomaly detection benchmark MVTec AD. As a result, compared to the strong baseline method GLASS, the proposed RePaste achieved improvements of 0.59$\%$ in the A2N and 0.50$\%$ in the N2A in terms of the S-AUROC metric. Furthermore, under conventional evaluation metrics such as AUROC and Per-Region Overlap (PRO), RePaste also achieved comparable performance to or even surpassed that of GLASS.


The main contributions of this paper are as follows.
\begin{itemize}
    \item We introduce “Anomaly-to-Normal scenario (A2N)” and “Normal-to-Anomaly scenario (N2A)”, which discover models that can easily update the definition of normal samples, thereby expanding the range of practical applications.
    \item We introduce “AUROC for Specification Changes”, which evaluates solely on the target of specification changes.
    \item We provide a systematic empirical study of existing anomaly detection methods under anomaly definition shifts and analyze their robustness.
    \item We introduce \textbf{RePaste}, a robust anomaly detection method for specification changes.
\end{itemize}

The structure of this paper is as follows. In Section \ref{sec:related}, we discuss related works. Section \ref{sec:method} explains the details of the proposed method. In Section \ref{sec:experiments}, we present experimental results and discussion. Finally, Section \ref{sec:conclusion} concludes our paper and describes future challenges.

\section{Related Works}
\label{sec:related}

We describe conventional anomaly detection methods from only normal samples. Since conventional methods are not designed to handle specification changes, the necessity of our scenarios is confirmed.

\textbf{Single-Class Anomaly Detection} is a framework in which an anomaly detection model is developed for each class (e.g., bottle, cable, etc.).
In recent years, five types of commonly used anomaly detection methods are memory-based \cite{padim,patchcore}, knowledge-distillation-based \cite{RD4AD,RDplus,ReconST}, flow-based \cite{fastflow,msflow,cflow}, reconstruction-based \cite{DDAD,anoVIT}, and pseudo-based methods \cite{cutpaste,memseg,simplenet,GLASS}.
Memory-based methods are simple approaches that embed images of normal samples into a compressed feature space. During inference, they compare the embedded features with those of an unknown sample, with anomalous samples expected to be more distant from the normal features.
Knowledge distillation-based methods use two models, a teacher model and a student model, and anomalies are detected based on the difference between their feature maps.
Flow-based methods use Normalizing Flow \cite{normalizing} to create a multidimensional Gaussian distribution, and features that deviate from this distribution are considered anomalies.
Reconstruction-based methods are trained only on normal images, so anomalous regions, being unseen during training, are poorly reconstructed, resulting in larger reconstruction errors used for anomaly detection.
Pseudo-anomaly-based methods involve adding artificially created anomalies to normal samples and then performing binary classification between normal and anomaly classes. These methods can be broadly divided into two types, those that add pseudo-anomalies to the input image and those that add them to the feature map.
In image-level approaches, anomalies are created by pasting random areas from normal images or texture datasets such as DTD \cite{dtd} onto random locations within the input image.
In feature-level approaches, Gaussian noise is added to the feature maps, and the resulting noisy features are treated as anomalies.
GLASS \cite{GLASS} which is the state-of-the-art method synthesizes controllable and distribution-aligned anomalies by applying gradient ascent and projected truncation to the Gaussian-noised feature maps.

\textbf{Multi-Class Anomaly Detection} aims to detect anomalies across multiple classes using a single unified model.
This approach eliminates the need to retrain separate models when new classes are introduced.
UniAD \cite{UniAD} proposed a unified reconstruction framework for anomaly detection.
UniNet \cite{UniNet} further extended this idea by introducing a unified model applicable to multiple domains.
Subsequent works include MambaAD \cite{mambaAD}, which leverages Mamba \cite{mamba} for superior long-range modeling and linear efficiency, DiAD \cite{DiAD} which utilizes the reconstruction capability of diffusion models, Dinomaly \cite{Dinomaly} which exploits the representational power of foundation models such as DINO \cite{DINO} and DINOv2 \cite{DINOv2}, and INP-Former \cite{INPFormer} which achieves anomaly detection by utilizing internal normal information within the image itself without referencing any external normal features.

However, all conventional methods are not designed to handle the definition changes of normal samples.
Therefore, it remains unclear which methods are robust to changes in the definition of normal.
To address this, we evaluate conventional methods on our proposed A2N and N2A scenarios.
In general, the specification changes in A2N and N2A scenarios are conducted in each class independently, 
so the methods specialized for Single-Class Anomaly Detection tend to achieve higher scores.

\section{Methodology}
\label{sec:method}

We aim to develop an anomaly detection model that can flexibly adapt to changes in the definition of normal samples. To this end, we propose two novel scenarios: “Anomaly-to-Normal scenario (A2N)” and “Normal-to-Anomaly scenario (N2A)” in Sec.\ref{sub:A2N} and \ref{sub:N2A}. Furthermore, to properly evaluate model performance under these scenarios, we introduce a novel metric called “AUROC for Specification Changes” in Sec.\ref{sub:EVAL}. In addition, we propose \textbf{RePaste} in Sec.\ref{sub:repaste}, a method designed to flexibly adapt to the definition changes of normal samples in our A2N and N2A scenarios.

\subsection{Anomaly-to-Normal Scenario (A2N)}
\label{sub:A2N}

\begin{figure}[t]
    \centering
    \includegraphics[width=1.0\linewidth]{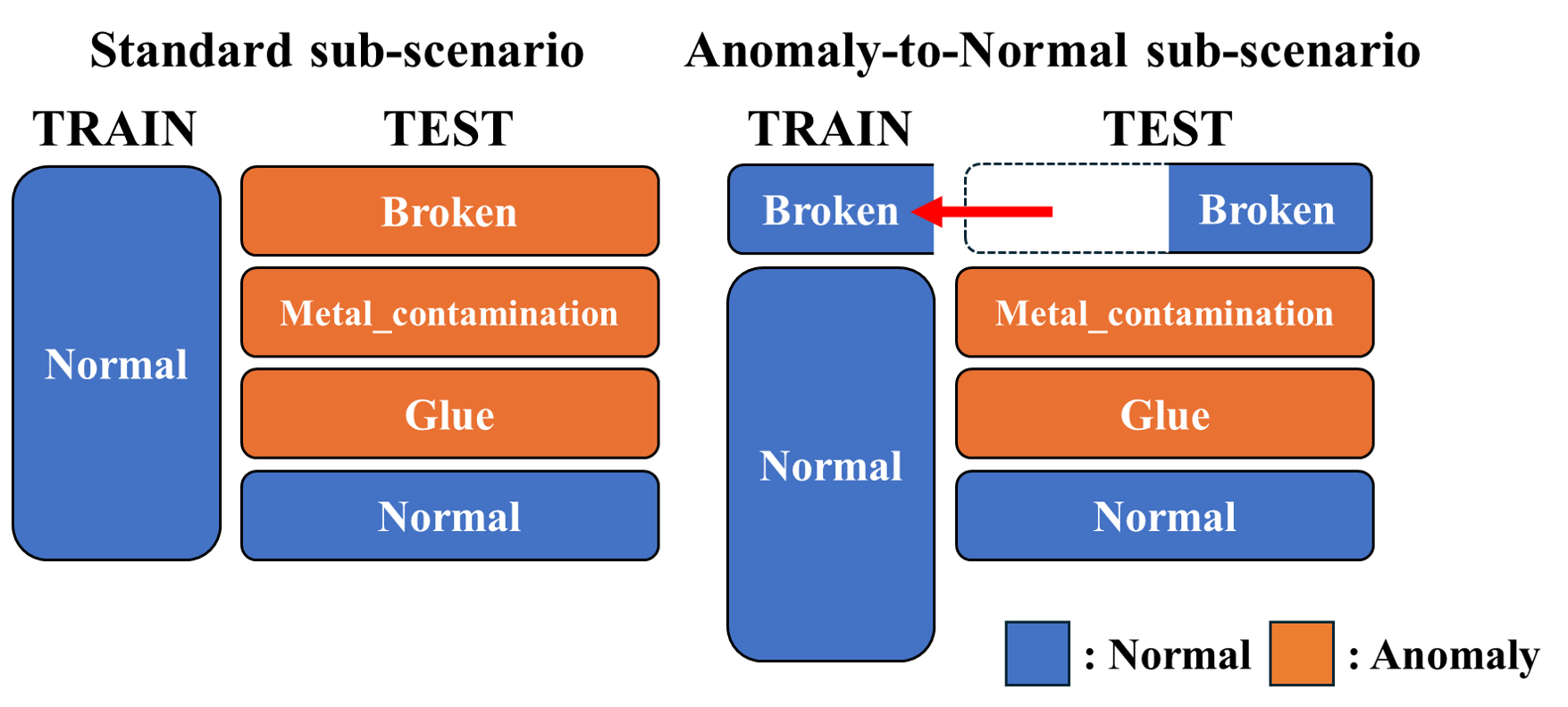}
    \caption{A2N includes “Anomaly-to-Normal sub-scenario" and “Standard sub-scenario". In the “Anomaly-to-Normal sub-scenario", if you want to treat the “Broken" as normal in the “Grid" category, we use half of all “Broken" images as training data and evaluate the remaining half as normal samples. “Standard sub-scenario" in A2N is the same as a standard anomaly detection task trained from normal samples. }
    \label{fig:A2N}
\end{figure}

\cref{fig:A2N} shows an example of a specification change in the “Grid” category, specifically for “Broken”, under the A2N. The A2N is divided into two sub-scenarios, namely the “Anomaly-to-Normal sub-scenario ($A2N_{A2N}$)” and the “Standard sub-scenario ($A2N_{S}$)”.
The purpose of creating two sub-scenarios is to specifically investigate how significant the differences would be when an anomalous sample is treated as a normal sample.
In the following sections, we explain the $A2N_{A2N}$ and $A2N_{S}$ by using the example of  specification change in the “Grid” category.

\subsubsection{Anomaly-to-Normal sub-scenario ($A2N_{A2N}$)}
\label{subsub:A2N}

In the $A2N_{A2N}$, one type of anomalous samples included in the test data, such as “Broken” is selected. The selected “Broken" images are split into two sets (e.g., if there are 40 images in total, they are split into 20 and 20). To treat the target of the specification change as normal samples, one half of the split images is added to the training data, and the remaining half is added to the test data.
If we include “Broken" in the training data, it will be treated as a normal sample. The reason for leaving the other half of the images in the test data is to evaluate how well it can be regarded as normal when we would like to treat “Broken" as a normal product. Normal samples originally included in the training data are denoted as $C_{train}$. 
Additionally, “Broken", which is the target for specification changes, is defined as $C_{t}^{A2N}$.
We define the training dataset $D_{train}^{A2N_{A2N}}$ as 
\begin{small}
\begin{eqnarray}
  D_{train}^{A2N_{A2N}} = \{ x \mid x \in C_{train}\} \cup \{ x \mid x \in C_{t}^{A2N}[:N/2]\}
  \label{equation:D_train_A2N_A2N}
\end{eqnarray}
\end{small}
where $N$ represents the total number of “Broken" images, and the first half of the images is used for training.

The test data includes both normal and anomalous samples. Specifically, the test data consist of normal samples, the remaining half of “Broken", “Metal$\_$contamination", and “Glue" as shown in \cref{fig:A2N}.  
“Metal$\_$contamination" and Glue" are anomalous samples, and they denoted as $C_{anomaly}^{A2N_{A2N}}$.
However, since “Broken" is designated as the target for specification changes, it is treated as normal samples.
\begin{eqnarray}
    \begin{aligned}
      D_{test}^{A2N_{A2N}} &= \{ x \mid x \in C_{test}\} \cup \{ x \mid x \in C_{t}^{A2N}[N/2:]\} \\
      &\cup \{ x \mid x \in C_{anomaly}^{A2N_{A2N}}\} 
    \end{aligned}
    \label{equation:D_test_A2N_A2N}
\end{eqnarray}
where $C_{test}$ is the original normal samples in the test data. Training is conducted on $D_{train}^{A2N}$, and evaluation is performed on $D_{test}^{A2N_{A2N}}$. 
This allows “Broken" to be treated as normal samples.

\subsubsection{Standard sub-scenario ($A2N_S$)}
\label{subsub:Standard_A2N}

Next, we explain the $A2N_S$ that is similar to the conventional anomaly detection from only normal samples. The training data consists solely of data defined as normal samples. We define the training dataset as 
$D_{train}^{A2N_S} = \{ x \mid x \in C_{train}\}$.

The test data includes both normal and anomalous samples. The categories of anomalous samples are defined as $C_{anomaly}^{A2N_S}$ which includes “Broken", “Metal$\_$contamination", and “Glue".
Test dataset $D_{test}^{A2N_S}$ can be represented as follows.
\begin{eqnarray}
  D_{test}^{A2N_S} = \{ x \mid x \in C_{test}\} \cup \{ x \mid x \in C_{anomaly}^{A2N_S}\}
  \label{equation:D_test_S}
\end{eqnarray}
Training is conducted on $D_{train}^{A2N_S}$, and evaluation is performed on $D_{test}^{A2N_S}$. Additionally, since “Broken", which is the subject of specification changes, is not included in $D_{train}^{A2N_S}$, it can be treated as anomalous samples in standard sub-scenario because we investigate the influence of the specification change.

In this case, we have used “Broken" as an example, but similar specification changes need to be made for all small anomaly types, 
including “Metal$\_$contamination" and “Glue". 
In \cref{sub:EVAL}, we explain evaluation metrics that reflect the specification changes when we treat anomalous samples as normal samples.

\subsection{Normal-to-Anomaly Scenario (N2A)}
\label{sub:N2A}

\begin{figure}[t]
    \centering
    \includegraphics[width=1.0\linewidth]{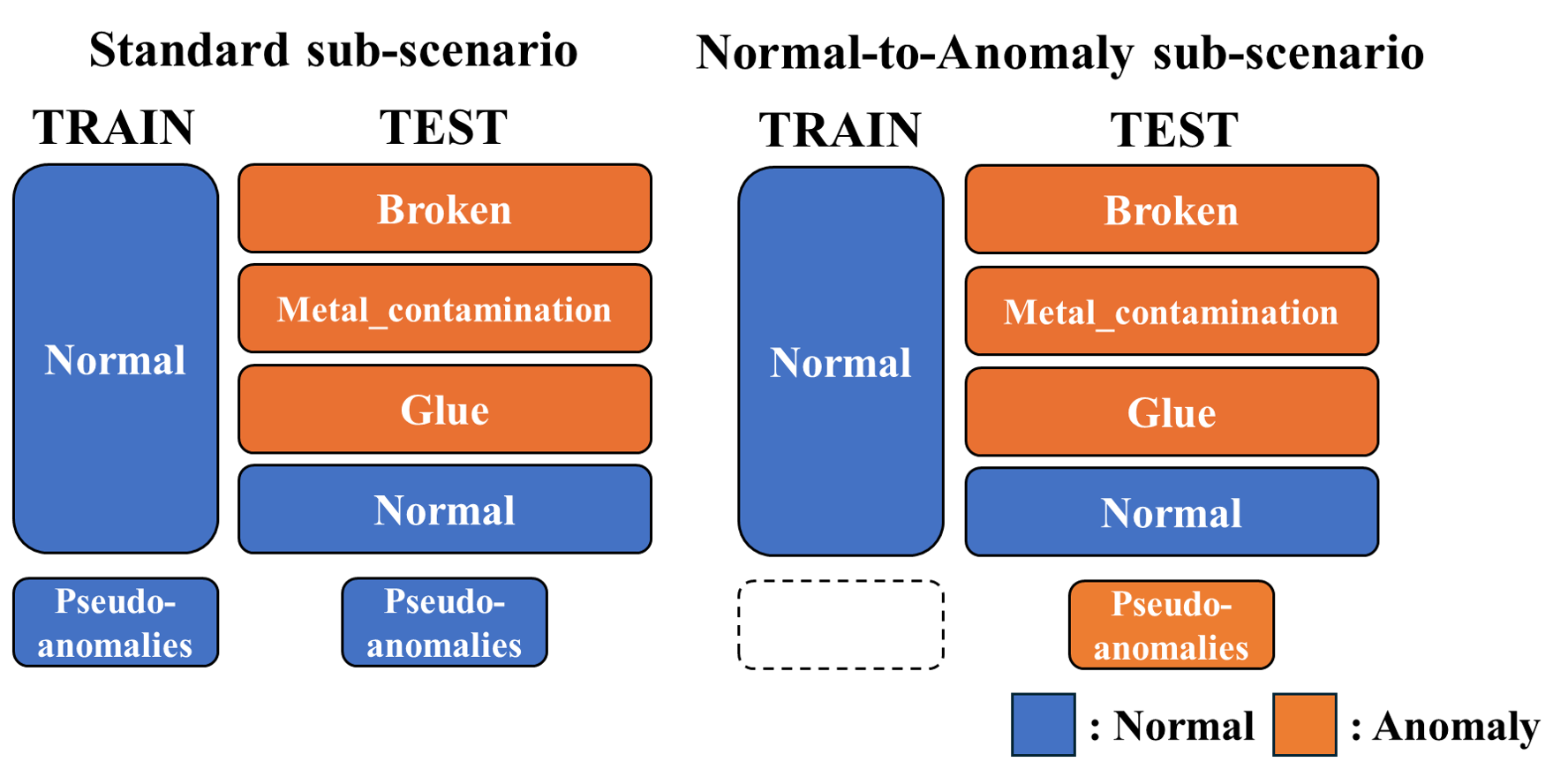}
    \caption{N2A includes “Normal-to-Anomaly sub-scenario" and “Standard sub-scenario". In the “Normal-to-Anomaly sub-scenario", pseudo-anomalies are added to the test data as anomalous samples. In “Standard sub-scenario", pseudo-anomalies are added to both training and test data as normal samples.
    }
    \label{fig:N2A}
\end{figure}

In \cref{fig:N2A}, the examples of “Grid" category in the N2A is shown.
The N2A assumes a situation where normal samples are reclassified as anomalies due to specification changes.
However, since the original training data in MVTec AD consist of only normal samples, it is difficult to directly use them for such specification changes.
To address this, pseudo-anomalies are used as substitutes for the targets of the specification changes. Pseudo-anomalies are generated by AnomalyAny \cite{anomalyany}, which can create diverse and realistic invisible anomalies. In this process, the pseudo-masks are generated by MemSeg \cite{memseg}.

N2A is divided into two sub-scenarios: the “Normal-to-Anomaly sub-scenario ($N2A_{N2A}$)” and the “Standard sub-scenario ($N2A_{S}$)”.
These sub-scenarios are designed to investigate how significantly the model’s performance changes when normal samples are treated as anomaly. Here we use pseudo-anomalies, however, in real-world situation, anomaly label is assigned to certain types of normal samples, and the N2A is evaluated.

\subsubsection{Normal-to-Anomaly sub-scenario ($N2A_{N2A}$)}
\label{subsub:N2A}

First, we explain the $N2A_{N2A}$ for the “Grid" category.
In this sub-scenario, pseudo-anomalies are added only to the test data, because they are treated as anomalous samples. In other words, the training dataset contains only the original normal samples. Thus, we can define the training dataset as $D_{train}^{N2A_{N2A}} = \{ x \mid x \in C_{train}\}$.
The test dataset includes both normal samples $C_{test}$ and anomalous samples. 
Specifically, the anomalous samples are “Broken", “Metal$\_$contamination", “Glue", and the pseudo-anomalies. Original anomalous test samples are referred to as $C_{anomaly}^{N2A_{N2A}}$, and the pseudo-anomalies, which are the targets for redefinition, are defined as $C_t^{N2A}$. 
Thus, we define test dataset as 
\begin{eqnarray}
    \begin{aligned}
      D_{test}^{N2A_{N2A}} &= \{ x \mid x \in C_{test}\} \cup \{ x \mid x \in C_{t}^{N2A}[N/2:]\} \\
      &\cup \{ x \mid x \in C_{anomaly}^{N2A_{N2A}}\}
    \end{aligned}
    \label{equation:D_test_N2A_N2A}
\end{eqnarray}
By training on $D_{train}^{N2A_{N2A}}$ and evaluating on $D_{test}^{N2A_{N2A}}$, we can evaluate the pseudo-anomalies as anomalous samples.

\subsubsection{Standard sub-scenario ($N2A_S$)}
\label{subsub:Standard_N2A}

The $N2A_S$ is simply adding pseudo-anomalies samples defined as normal samples to $D_{train}^{N2A}$ in the N2A as shown in \cref{fig:N2A}. We define this training dataset as
\begin{eqnarray}
  D_{train}^{N2A_S} = \{ x \mid x \in C_{train}\} \cup \{ x \mid x \in C_t^{N2A}[:N/2]\}
  \label{equation:D_train_N2A_S}
\end{eqnarray}
The test dataset is exactly the same as Equation \ref{equation:D_test_N2A_N2A}. Thus, $D_{test}^{N2A_S} = D_{test}^{N2A_{N2A}}$. However, since pseudo-anomalies are included in the training data, they are treated as normal samples not anomalous samples. 
Training is performed on $D_{train}^{N2A_S}$, and evaluation is conducted on $D_{test}^{N2A_S}$. This allows the evaluation of pseudo-anomalies as normal samples.

In $N2A_{N2A}$ and $N2A_S$, pseudo-anomalies are treated as anomalous samples and normal samples, respectively. By using these two sub-scenarios, it is possible to evaluate the degree of the difference that appears when normal samples are treated as anomalous samples.

\subsection{AUROC for Specification Changes}
\label{sub:EVAL}

Conventional evaluation metrics, such as AUROC and F1-score, assume a fixed definition of normal and anomalous samples and therefore can not quantify a model’s adaptability to changes in these definitions. To address this problem, we propose AUROC for Specification Changes (S-AUROC), a metric that focuses on samples affected by specification changes, enabling the evaluation of how flexibly a model adapts to the evolving definitions of normal and abnormal.


By using the A2N presented in \cref{sub:A2N} as an example, we assume a situation in which the anomaly class “Broken” is redefined as normal.
We first prepare two models trained under different sub-scenarios: the Standard sub-scenario, where “Broken” is treated as anomalous, and the Anomaly-to-Normal sub-scenario, where “Broken” is regarded as normal.
Next, the same set of “Broken” images is fed into both models to obtain their corresponding anomaly maps.
Under the Standard sub-scenario, “Broken” samples are considered anomalies, whereas under the Anomaly-to-Normal sub-scenario, they are treated as normal samples.
Based on these outputs, we compute the AUROC for each sub-scenario using the respective normal and abnormal definitions. By comparing these AUROC scores, we evaluate how flexibly the model adapts to the redefinition of abnormal to normal.

\subsection{RePaste}
\label{sub:repaste}

\begin{figure}[t]
    \centering
    \includegraphics[width=0.85\linewidth]{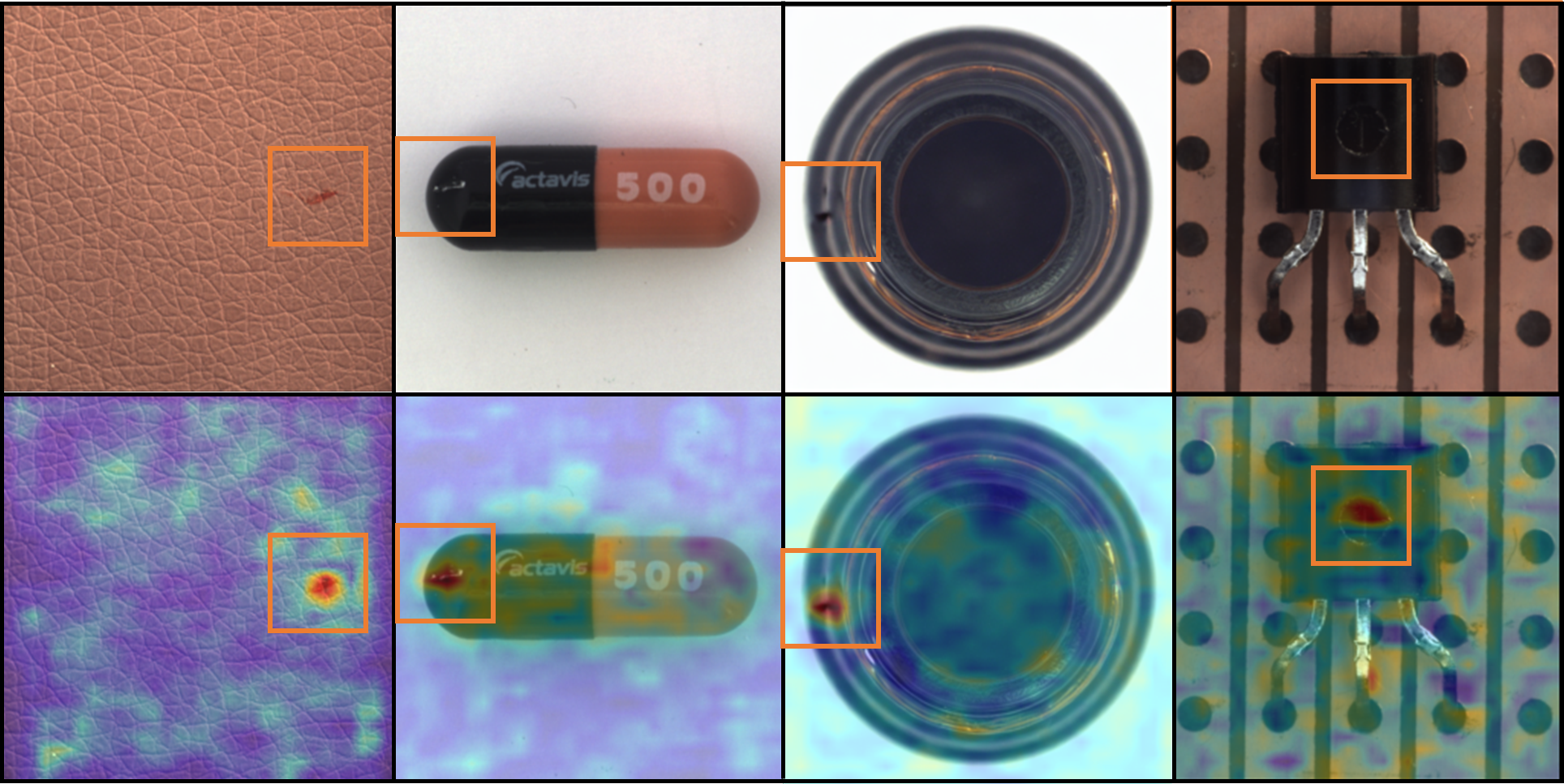}
    \caption{Examples where the anomaly score map shows a high area in the \textcolor{orange}{orange} regions, even though they are used as normal samples during training. Note that we use GLASS.}
    \label{fig:ob}
\end{figure}

When manufacturing specifications change, certain regions that were previously considered anomalous, such as small scratches or dust, may need to be redefined as normal. However, conventional anomaly detection models tend to assign persistently high anomaly scores to such regions, causing them to be repeatedly detected as false positives even after specification updates.
To address this problem and enable flexible redefinition of normal samples, we propose a training-time augmentation strategy called \textbf{RePaste}.

\begin{figure}[t]
    \centering
    \includegraphics[width=0.75\linewidth]{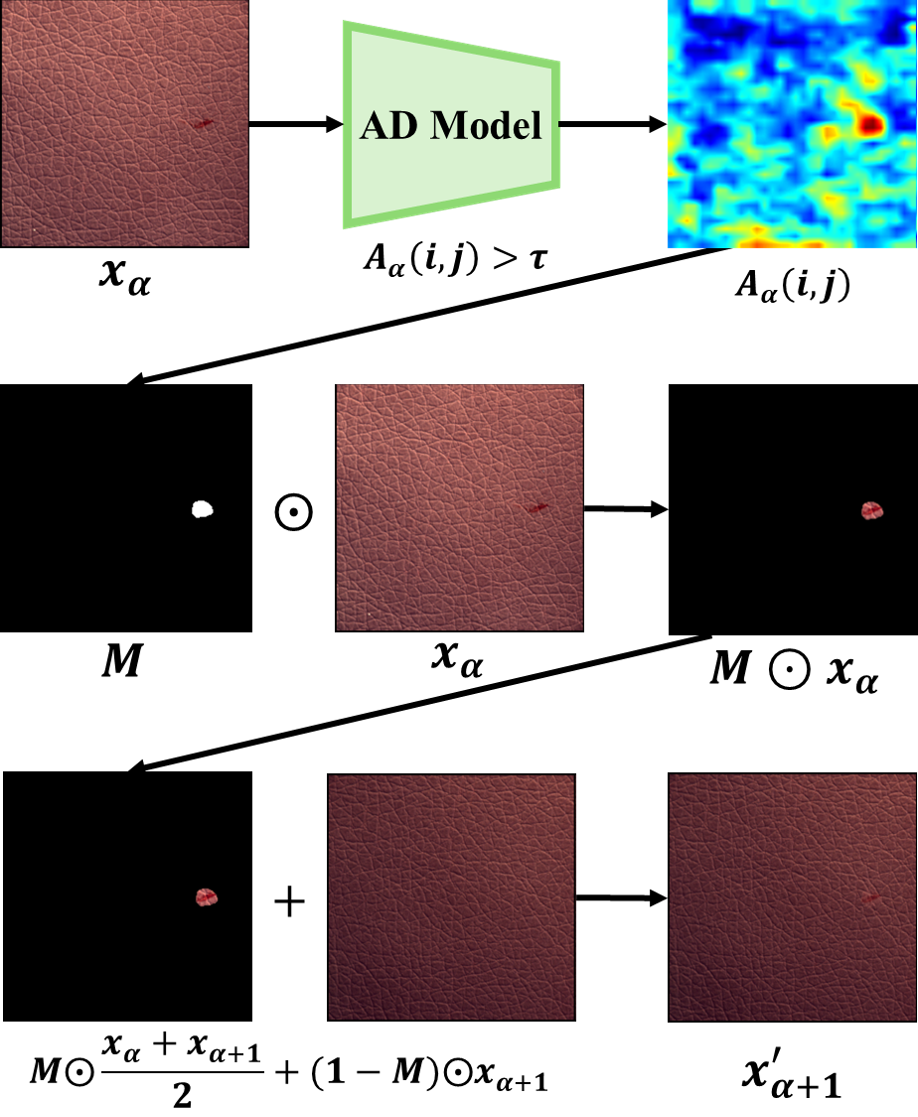}
    \caption{Overview of Repaste. To detect small scratches and dust that are often subject to specification changes, RePaste feeds the input image $x_\alpha$ into the model to obtain an anomaly map $A_\alpha$. Then, to extract only the regions with high anomaly scores, $A_\alpha$ is binarized using a threshold $\tau$. Finally, the extracted regions containing small scratches or dust are re-pasted onto the next image $x_{\alpha+1}$, increasing their occurrence frequency and making them easier to be treated as normal samples.}
    \label{fig:RePaste_overview}
\end{figure}

The motivation RePaste arises from the observation that regions exhibiting consistently high anomaly scores during training often correspond to visually minor defects, including dust or small scratches (\cref{fig:ob}). Importantly, the normal and anomalous status of these regions is highly sensitive to specification changes, making them a primary source of false positives after such updates. Based on this observation, RePaste aims to explicitly reduce the anomaly scores of these regions by increasing their occurrence frequency during training, encouraging the model to incorporate them into the normal features under updated specifications.

\cref{fig:RePaste_overview} illustrates the overview of RePaste. During training, the input image $x_\alpha \in \mathbb{R}^{3 \times H \times W}$ is fed into the model, and we obtain the anomaly map $A_\alpha \in \mathbb{R}^{H \times W}$ where $\alpha$ denotes the current training iteration, $A_\alpha$ is resized to $H \times W$ by bilinear interpolation. Next, pixels with anomaly scores exceeding a threshold $\tau$ are regarded as anomalous regions, and a mask $M \in \mathbb{R}^{H \times W}$ is generated as 
\begin{equation}
M(i, j) =
\begin{cases}
1, & \text{if } A_\alpha(i, j) > \tau, \\
0, & \text{otherwise.}
\end{cases}
\label{eq:mask}
\end{equation}
Using this mask, the high score regions from the current image $x_\alpha$ are re-pasted onto the next input image $x_{\alpha+1}$, which is sampled from the original set of normal training images, to generate a new training image. At this stage, to mitigate the discontinuities that arise at the pasted boundaries, we draw inspiration from Mixup \cite{mixup} and construct the boundaries to be as natural as possible.
\begin{equation}
x'_{\alpha+1} = M \odot \frac{x_\alpha + x_{\alpha+1}}{2} + (1 - M) \odot x_{\alpha+1}
\label{eq:mixup}
\end{equation}
RePaste does not increase the size of the training set. Instead, $x_{\alpha+1}^{'}$ replace $x_{\alpha+1}$ for the current training iteration.
During inference, the input image $x \in \mathbb{R}^{3 \times H \times W}$ is fed into the model, and an anomaly score $A$ is obtained. Thus, RePaste is not required, and the inference time is the same as that of conventional anomaly detection models.

Overall, RePaste provides a simple yet effective mechanism to adapt anomaly detection models to specification changes without modifying the model architecture or inference procedure.
By selectively increasing the exposure of regions that are likely to be redefined as normal, the model gradually suppresses their anomaly responses, while simultaneously reducing false positives caused by background artifacts such as dust, and preserving sensitivity to other anomaly types.
Importantly, RePaste is applied only during training and does not introduce any additional computational cost at inference time, making it practical for real-world deployment under evolving specifications.


\section{Experiments}
\label{sec:experiments}

\subsection{Experimental Details}

\textbf{Datasets.}
To evaluate anomaly detection performance in the industrial domain, our experiments were conducted on the MVTec AD benchmark \cite{mvtec}.
MVTec AD consists of 15 classes such as cable and capsule
with a total of 5,354 images, of which 1,725 belong to the test set.
Each class is divided into training data which contain only normal samples and test data which contain both normal and anomalous samples. Each class corresponds to a product and includes various defect types, along with the corresponding ground-truth anomaly masks.
For the methods using WideResNet \cite{wide}, the input images are resized to $256 \times 256$ by following \cite{patchcore,simplenet}.
For the methods using DINOv2 \cite{DINOv2}, the images are resized to $252 \times 252$, as it only accepts the dimensions that are multiples of 14.

\textbf{A2N and N2A settings.}
In $A2N_{A2N}$, as shown in \cref{fig:A2N}, “Broken” is the target of specification change, and half of the images in the “Broken” are redefined as normal samples and included in training data. In A2N, since small defects such as scratches or dust are the targets of specification change, large anomalies are not subject to the change. Specifically, anomalies whose average size in the ground-truth masks is less than 1$\%$ of the image area are defined as targets with small anomalies for specification change.
In $N2A_S$, 40 pseudo-anomaly images are generated using AnomalyAny \cite{anomalyany} and MemSeg \cite{memseg}. As shown in \cref{fig:N2A}, 20 images are used for training and the remaining 20 are used for evaluation.
In contrast, in $N2A_{N2A}$, since the pseudo-anomaly images are treated as anomalies according to the specification change, the pseudo-anomaly images included in the training data are removed.

\textbf{Evaluation Metrics.}
The samples for specification change in A2N and N2A are evaluated by AUROC for Specification Changes (S-AUROC), explained in \cref{sub:EVAL}.
In addition, to evaluate the capability of distinguishing between normal and anomalous samples, we use the same Image-level AUROC (I-AUROC), Pixel-level AUROC (P-AUROC), and Per-Region Overlap (PRO) for evaluation as conventional methods \cite{RDplus}.

\textbf{Implementation Details.}
As the baseline, we use GLASS \cite{GLASS} which achieved state-of-the-art results, and integrate our proposed RePaste into it.
Therefore, the experimental settings are the same as those of GLASS. In our RePaste, the value of $\tau$ is set to 0.9.

\subsection{Comparison of State of the Art on MVTec AD}

\begin{table*}[t]
    \centering
    \caption{Comparison results for the proposed scenarios on MVTec AD. The evaluation metric is the S-AUROC scores. 
    }
    \scalebox{0.59}{
    \begin{tabular}{l||cccccccccc||cc}
        \hline
        Model & FastFlow \cite{fastflow} & PatchCore \cite{patchcore} & RD4AD \cite{RD4AD} & RD++ \cite{RDplus} & SimpleNet \cite{simplenet} & DiAD \cite{DiAD}  & mambaAD \cite{mamba} & INP-Former \cite{INPFormer} & UniNet \cite{UniNet} & Dinomaly \cite{Dinomaly} & \cellcolor{red!10}GLASS \cite{GLASS} & \cellcolor{red!10}RePaste \\
        Conf & Arxiv2021 & CVPR2022 & CVPR2022 & CVPR2023 & CVPR2023 & AAAI2024  & NeurIPS2024 & CVPR2025 & CVPR2025 & CVPR2025 & \cellcolor{red!10}ECCV2024 & \cellcolor{red!10}- \\
        \hline
        A2N & 82.11 & 50.75 & 65.26 & 67.68 & 84.25 & 54.33 & 62.39 & 68.69 & 61.34 & 84.70 & \cellcolor{red!10}\underline{86.29} & \cellcolor{red!10}\textbf{86.88} \\
        N2A & 79.68 & 50.23 & 72.70 & 77.20 & 75.70 & 52.48 & 58.82 & 60.47 & 72.87 & 81.88 & \cellcolor{red!10}\underline{83.25} & \cellcolor{red!10}\textbf{83.75}  \\
        \hline
    \end{tabular}
    }
    \label{tab:scenario}
\end{table*}

\cref{tab:scenario} shows the evaluation results of many anomaly detection methods for specification change on MVTec AD. 
The results demonstrated that our method is the most robust to specification changes. 
Among the conventional methods, GLASS achieved the highest total score when combining A2N and N2A. We consider that this is because GLASS generates anomalies using noise, and through gradient ascent, it produces better anomalies, allowing it to flexibly adapt to specification changes.
PatchCore performed almost like random guessing. This is because coreset sampling often removes rare features, so features related to specification changes are not well learned. As a result, their distances remain large during inference, causing misclassification.
Our proposed RePaste based on GLASS achieved the best performance compared to conventional anomaly detection models. Furthermore, compared with the GLASS, RePaste improved S-AUROC by 0.59$\%$ in A2N and 0.50$\%$ in N2A. 
In summary, RePaste represents an important and meaningful improvement in realistic industrial inspection settings.

\begin{table*}[t]
    \centering
    \caption{Comparison results of $A2N_S$ , $A2N_{A2N}$, $N2A_S$ and $N2A_{N2A}$ on MVTec AD. 
    The evaluation metric is the mean I-AUROC, P-AUROC, and PRO. 
    }
    \scalebox{0.67}{
    \begin{tabular}{c|l|cccccccccc||cc}
    \hline
     Eval& Scenario & FastFlow & PatchCore & RD4AD & RD++ & SimpleNet & DiAD & mambaAD & INP-Former & UniNet & Dinomaly & \cellcolor{red!10}GLASS & \cellcolor{red!10}RePaste \\
    \hline

    \multirow{4}{*}{I-AUROC}
    & $A2N_S$ 
    & 96.02 & 98.33 & 96.97 & 97.09 & 98.60 & 84.40 & 98.63 & 96.65 & 96.07 & 98.49 & \cellcolor{red!10}\textbf{99.54} & \cellcolor{red!10}\underline{99.45} \\
    
    & $A2N_{A2N}$ 
    & 91.05 & 91.31 & 89.07 & 89.16 & 93.47 & 77.06 & 90.13 & 92.37 & 87.68 & 93.74 & \cellcolor{red!10}\underline{94.65} & \cellcolor{red!10}\textbf{94.77} \\
    
    & $N2A_S$ 
    & 93.46 & 93.67 & 90.47 & 89.88 & 94.24 & 85.42 & 94.92 & 95.20 & 88.80 & \textbf{96.28} & \cellcolor{red!10}\underline{95.68} & \cellcolor{red!10}\underline{95.68} \\
    
    & $N2A_{N2A}$ 
    & 96.16 & 95.06 & 95.11 & 95.36 & 97.29 & 82.83 & 95.32 & \underline{97.79} & 94.92 & \textbf{98.24} & \cellcolor{red!10}97.45 & \cellcolor{red!10}97.61 \\
    \hline
    \multirow{4}{*}{P-AUROC}
    & $A2N_S$ 
    & 98.60 & 99.10 & 97.12 & 97.01 & 98.85 & 93.87 & 99.13 & 98.75 & 98.81 & 99.08 & \cellcolor{red!10}\textbf{99.37} & \cellcolor{red!10}\underline{99.33} \\
    
    & $A2N_{A2N}$ 
    & 98.58 & 99.12 & 95.70 & 96.46 & 98.86 & 94.45 & 99.06 & 98.86 & 98.75 & 99.11 & \cellcolor{red!10}\textbf{99.31} & \cellcolor{red!10}\textbf{99.31} \\
    
    & $N2A_S$ 
    & 98.20 & 98.52 & 94.25 & 96.13 & 98.30 & 93.74 & 97.88 & 97.57 & 98.52 & \textbf{99.08} & \cellcolor{red!10}98.09 & \cellcolor{red!10}\underline{99.06} \\
    
    & $N2A_{N2A}$ 
    & 96.92 & 97.78 & 93.51 & 93.78 & 97.50 & 92.85 & 97.12 & 96.11 & 97.74 & \textbf{98.34} & \cellcolor{red!10}97.43 & \cellcolor{red!10}\underline{98.28} \\
    \hline
    \multirow{4}{*}{PRO}
    & $A2N_S$ 
    & 94.73 & 96.35 & 92.15 & 91.84 & 94.67 & 80.42 & \underline{97.33} & 97.08 & 96.15 & 96.14 & \cellcolor{red!10}96.95 & \cellcolor{red!10}\textbf{97.45} \\
    
    & $A2N_{A2N}$ 
    & 94.45 & 96.39 & 88.52 & 90.19 & 94.50 & 81.46 & 96.82 & \textbf{96.86} & 96.23 & 96.08 & \cellcolor{red!10}\underline{96.85} & \cellcolor{red!10}\underline{96.85} \\
    
    & $N2A_S$ 
    & 94.03 & 95.10 & 87.77 & 87.43 & 92.96 & 83.01 & 94.82 & 94.26 & 92.52 & 95.13 & \cellcolor{red!10}\textbf{96.43} & \cellcolor{red!10}\underline{96.24} \\
    
    & $N2A_{N2A}$ 
    & 85.25 & 87.99 & 79.49 & 79.07 & 85.43 & 75.16 & 87.87 & 87.91 & 84.66 & 
    88.72 & \cellcolor{red!10}\underline{90.16} & \cellcolor{red!10}\textbf{90.24} \\
    \hline
    Mean &
    & 94.79 & 95.73 & 91.68 & 91.95 & 95.39 & 85.39 & 95.75 & 95.78 & 94.24 & 96.54 & \cellcolor{red!10}\underline{96.83} & \cellcolor{red!10}\textbf{97.02} \\
    \hline
    \end{tabular}
    }
    \label{tab:A2N}
\end{table*}

We demonstrated that RePaste achieves State of the Art (SOTA) performance for the samples of specification changes.
However, the performance on the entire dataset, including both specification-change and non-specification-change samples, is also important for anomaly detection. 
If the performance of the overall system dropped after specification changes, we can not use the method in real environment. 
Thus, we evaluate the performance by AUROC and PRO, which is used in standard anomaly detection.

\cref{tab:A2N} shows the mean I-AUROC, P-AUROC, and PRO for the overall, in the $A2N_S$, $A2N_{A2N}$, $N2A_S$, and $N2A_{N2A}$ scenarios. 
RePaste achieves performance that is comparable to or better than existing methods across all evaluation metrics (I-AUROC, P-AUROC, and PRO) and all scenarios. In particular, RePaste attains the highest Mean score of 97.02 $\%$ among all methods, quantitatively demonstrating its effectiveness for anomaly detection.

Focusing first on the A2N, RePaste shows performance almost equivalent to GLASS. Notably, for the PRO metric in the $A2N_S$, RePaste outperforms GLASS by 0.5$\%$. This improvement can be attributed to the model’s ability to successfully treat background artifacts such as dust or minor texture variations previously causing false positives as normal features through re-pasting during training.

Next, in the N2A, the advantage of RePaste becomes particularly evident in terms of P-AUROC. Compared to GLASS, RePaste achieves improvements of 0.97$\%$ in $N2A_S$ and 0.85$\%$ in $N2A_{N2A}$. This can be explained by the re-pasting, which enables the model to suppress regions that would otherwise be falsely detected as anomalies when pseudo-anomalous regions are intended to be regarded as normal after specification updates. Conversely, when pseudo-anomalous regions should be treated as true anomalies, RePaste still effectively reduces background-induced false positives, consistent with the observations in the A2N.

Importantly, RePaste requires no additional annotations and introduces no extra processing at inference time, functioning solely as a simple training-time data augmentation strategy. Despite this simplicity, it consistently matches or surpasses the performance of strong baselines including GLASS. These results highlight the practical applicability of RePaste in real-world industrial anomaly detection settings, where frequent specification changes are inevitable.

\begin{figure}[t]
    \centering
    \includegraphics[width=1.0\linewidth]{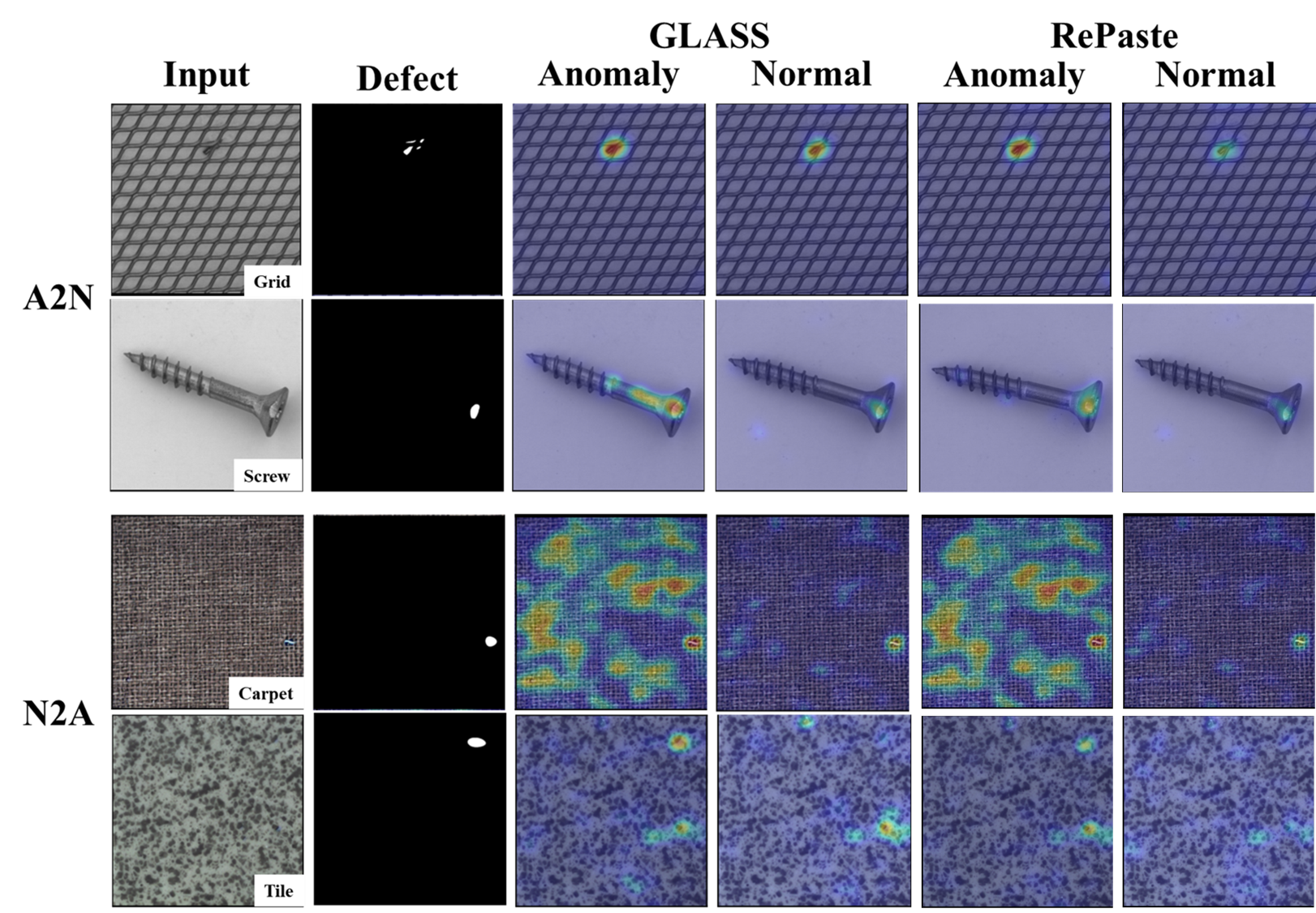}
    \caption{Qualitative visualization for pixel-level anomaly segmentation.}
    \label{fig:visual}
\end{figure}

\subsection{Qualitative Results.}

\cref{fig:visual} shows qualitative results in A2N and N2A scenarios on the MVTec AD, and confirmed that RePaste demonstrates high performance in anomaly segmentation.
In A2N, when an image is intended to be treated as anomalous, both GLASS and RePaste correctly identify it as an anomaly.
However, when it is intended to be treated as normal, RePaste correctly recognizes it as normal with small anomaly score, whereas GLASS still assigns high anomaly scores to the defective regions.

In contrast, in the Carpet category of N2A, both GLASS and RePaste tend to classify the entire image as anomalous. We consider that this behavior is due to the model recognizing that the image is generated, probably because its texture differs from that of real-world images.
On the other hand, in the Tile category of N2A, when an image is intended to be treated as anomalous, it is correctly detected as an anomaly. Moreover, when it is intended to be treated as normal, RePaste successfully recognizes it as normal, whereas GLASS still classifies it as anomalous.


\subsection{Abulation study}
\label{sub:abu}

\begin{table}[t]
    \centering
    \caption{Comparison of GLASS, RePaste w/ and w/o Mixup.
    }
    \scalebox{0.72}{
        \begin{tabular}{lccc}
        \hline
        & GLASS & RePaste & RePaste \\ 
        \hline
        Mixup&  &  & \checkmark \\ 
        \hline
        \multicolumn{4}{c}{S-AUROC} \\ 
        \hline
        $A2N$ & 86.29 & \textbf{87.48} & 86.88 \\
        $N2A$ & 83.25 & 78.26 & \textbf{83.75} \\
        \hline
        \multicolumn{4}{c}{I-AUROC} \\ 
        \hline
        $A2N_S$ & \textbf{99.54} & 99.53 & 99.45 \\
        $A2N_{A2N}$ & 94.65 & 94.17 & \textbf{94.77} \\
        $N2A_S$ & \textbf{95.68} & 95.58 & 98.67 \\
        $N2A_{N2A}$ & 97.45 & 97.50 & \textbf{97.61} \\
        \hline
        
    \end{tabular}}
    \label{tab:Mixup}
\end{table}

\textbf{Image gap caused by re-pasting}
Our proposed RePaste suppresses boundary effects by drawing inspiration from Mixup. However, it remains unclear whether this suppression influences anomaly detection performance. Therefore, we conducted experiments by modifying \cref{eq:mixup} as follows.
Specifically, we modify \cref{eq:repaste} as 
\begin{equation}
x'_{t+1} = M \odot x_t + (1 - M) \odot x_{t+1}.
\label{eq:repaste}
\end{equation}

\begin{figure}[t]
    \centering
    \includegraphics[width=0.75\linewidth]{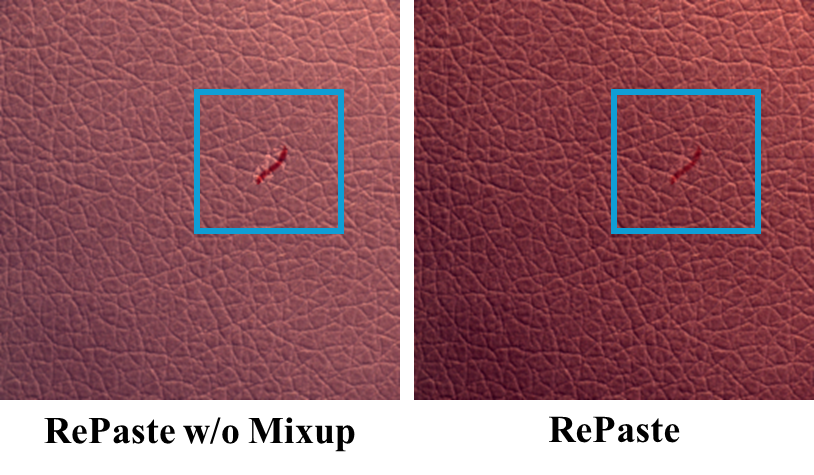}
    \caption{The difference between RePaste with and without Mixup.
        The blue box regions show the RePaste areas.
        RePaste Without Mixup creates boundary discontinuities.}
    \label{fig:withmixup_visual}
\end{figure}

\cref{tab:Mixup} shows a comparison of RePaste with and without the use of Mixup. We employ S-AUROC and I-AUROC as evaluation metrics.
From the results, under the S-AUROC metric, RePaste without Mixup achieved 0.6$\%$ performance improvement over RePaste in the A2N scenario. However, in the N2A scenario, RePaste without Mixup exhibits 5.49$\%$ performance degradation compared to RePaste. 
We attribute this degradation to boundary discontinuities, which, as shown in \cref{fig:withmixup_visual}, disturb the feature distribution and consequently degrade performance.
In contrast, RePaste consistently outperforms GLASS with more stable performance. This suggests that RePaste effectively minimizes boundary effects while re-pasting small anomalies such as dust, thereby reducing false positives.
Next, under the I-AUROC metric in the $A2N_{A2N}$ setting, RePaste without Mixup shows 0.6$\%$ performance drop compared to RePaste with Mixup. This degradation can be explained by the same reason: boundary discontinuities lead to instability in the feature distribution. 
On the other hand, RePaste achieved  comparable performance to or better than GLASS. Therefore, we conclude that incorporating Mixup into RePaste is beneficial.

\section{Conclusion}
\label{sec:conclusion}

We proposed novel scenarios, evaluation metrics, and RePaste to address the issue of ambiguity in the definition of normal samples in anomaly detection. 
As a result, it achieved SOTA performance in both A2N and N2A scenarios under the proposed S-AUROC which is the metric for Specification Changes.
Furthermore, RePaste demonstrated comparable performance or superior to GLASS which is the best method among conventional methods, in terms of AUROC and PRO.
We hope that the proposed scenarios and the S-AUROC will facilitate the development of more robust methods for handling specification changes.
{
    \small
    \bibliographystyle{ieeenat_fullname}
    \bibliography{main}
}


\end{document}